\documentclass[10pt,twocolumn,letterpaper]{article}
\usepackage{wacv}
\usepackage{times}
\usepackage{epsfig}
\usepackage{graphicx}
\usepackage{amsmath}
\usepackage{amssymb}
\usepackage{float}
\usepackage{caption}
\usepackage{booktabs}

\def\wacvPaperID{1174} % Enter the WACV Paper ID here

\wacvfinalcopy % *** Uncomment this line for the final submission

\ifwacvfinal
\def\assignedStartPage{1} % *** Enter the assigned starting page number (instead of 9876)
\fi

\ifwacvfinal
\usepackage[breaklinks=true,bookmarks=false]{hyperref}
\else
\usepackage[pagebackref=true,breaklinks=true,colorlinks,bookmarks=false]{hyperref}
\fi

\ifwacvfinal
\else
\fi

\begin{document}

%%%%%%%%% TITLE
\title{MixerGAN: An MLP-Based Architecture for \\Unpaired Image-to-Image Translation}

\author{George Cazenavette\\
Robotics Institute\\
Carnegie Mellon University\\
{\tt\small gcazenav@cs.cmu.edu}
\and
Manuel Ladron De Guevara\\
School of Architecture\\
Carnegie Mellon University\\
{\tt\small manuelr@andrew.cmu.edu}
}

\twocolumn[{
\maketitle
\begin{center}
    \includegraphics[width=1.0\linewidth]{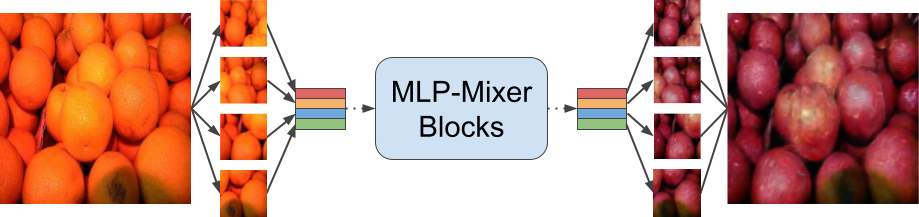}
    \captionof{figure}{Overview of MixerGAN. Our generator linearly projects image patches into vectors before translating via mixer blocks and up-projecting to transfigured patches.}
    \label{fig:my_label}
\end{center}
}]
%\thispagestyle{empty}

%%%%%%%%% ABSTRACT
\begin{abstract}
   While attention-based transformer networks achieve unparalleled success in nearly all language tasks, the large number of tokens (pixels) found in images coupled with the quadratic activation memory usage makes them prohibitive for problems in computer vision. As such, while language-to-language translation has been revolutionized by the transformer model, convolutional networks remain the de facto solution for image-to-image translation. The recently proposed MLP-Mixer architecture alleviates some of the computational issues associated with attention-based networks while still retaining the long-range connections that make transformer models desirable. Leveraging this memory-efficient alternative to self-attention, we propose a new exploratory model in unpaired image-to-image translation called MixerGAN: a simpler MLP-based architecture that considers long-distance relationships between pixels without the need for expensive attention mechanisms. Quantitative and qualitative analysis shows that MixerGAN achieves competitive results when compared to prior convolutional-based methods.
\end{abstract}

%%%%%%%%% BODY TEXT
\section{Introduction}
As with most visual tasks in the deep learning age, image-to-image translation has been dominated by fully convolutional neural networks since its inception \cite{pix2pix, cyclegan, cut}. However, the recently proposed MLP-Mixer \cite{mixer} architecture showed that simple multi-layered perceptrons are still useful for visual learning in deep neural networks. While the original MLP-Mixer paper only addressed \textit{discriminative} tasks, we propose adapting this new architecture for \textit{generative} tasks. Since the MLP-Mixer closely resembles the transformer model originally used for sequence-to-sequence translation in natural language processing \cite{attention}, we choose to evaluate the MLP-Mixer's capabilities on image-to-image translation through our new model: MixerGAN.

Transformer models are desirable for sequence-to-sequence translation because their self-attention mechanisms account for long-range dependencies that are overlooked by the relatively small receptive fields of convolutions. As such, self-attention has been used in image-to-image translation before \cite{tang2019attentiongan, NEURIPS2018_mejjati, emami2020spa}. However, the extreme memory usage of self-attention mechanisms prevents them from being used effectively on whole images, even after they have been significantly down-sampled. Motivated by a memory-efficient alternative to self-attention mechanisms, MixerGAN offers a much cheaper alternative to transformers that still accounts for long-range dependencies within an image.

Image-to-image translation is the task of learning a mapping between a source domain and a target domain. This is useful for applications such as data augmentation \cite{mariani2018dataaugm}, domain adaptation \cite{murez2018adaptation}, colorization \cite{cao2017colorization}, style transfer \cite{pix2pix}, or image super-resolution \cite{ledig2017superresolution, wu2017srpgan}. In an ideal setting, we have source-target paired inputs and this problem can be solved using supervised learning. This is the setting of the seminal work from Isola et al. \cite{pix2pix} where an encoder-decoder-like network translates from one domain $X$ to another $Y$ and vice-versa. However, finding paired data is hard, and the lack of such datasets led to the development of unsupervised approaches like CycleGAN \cite{cyclegan}, where a cycle consistency loss ensures an invertible mapping between the two domains. 

Although we show that MixerGAN achieves results competitive with prior convolutional-based generative models, our goal is not to achieve state-of-the-art results or out-perform any benchmarks on image-to-image translation. Instead, we establish a new baseline that is able to provide similar results as those in \cite{cyclegan}, showing that an MLP-based architecture can be used for image synthesis tasks. 

In short, our primary contribution is showing that the MLP-Mixer architecture can be adapted to effectively perform unpaired image-to-image translation while accounting for long-range dependencies at a much lower cost than a transformer model.

% Attention mechanisms \cite{bahdanau2014neural} and the proposed Transformer model \cite{attention} revolutionized sequence-to-sequence language models due to the attention's ability to focus on relevant parts of an input sequence, solving the deficit of long range interactions from recurrent neural networks, improving state-of-the-art results in a wide range of natural language processing tasks. Attention-based models became the de facto neural model in the language community. Inspired by this, great efforts have been done trying to combine convolutional neural networks (CNNs) with self-attention \cite{wang2018non, carion2020end}, or replacing convolutions entirely with a stand-alone self-attention model for discriminative tasks. Dosovitskiy et al. introduced the visual transformer (ViT) \cite{ViT}, slicing the input image into a sequence of patches to decrease the input sequence size, showing competitive results for image classification tasks. However, these models do not scale well mostly due to the attention's quadratic cost with respect to the input and thus, CNNs architectures are still the backbone for many large-scale computer vision applications.

% The new MLP-mixer architecture \cite{mixer} has been proposed as a less memory-intensive alternative to transformer models for visual discriminative tasks.

\section{Related Work}

Throughout the evolution of sequence-to-sequence translation, it has been shown that context-awareness is critical to high-efficacy models \cite{elmo}. Traditional convolutional models limit context sharing between tokens to the width of the filter, preventing the model from learning longer-range relations between distant tokens. To that end, the natural language processing community moved on from these traditional models and adopted transformer models \cite{attention} which offer connections between all tokens in the sequence while also eliminating the inductive locality bias. In this section, we first briefly review general CNN-based image-to-image translation frameworks, followed by models that propose different reinterpretations of attention mechanisms to address longer range and better informed dependencies in pixel space. 

\paragraph{Image-to-image translation} GAN \cite{goodfellow2014generative} frameworks are the natural and standard approaches to image-to-image translation tasks. For instance, Isola et al. \cite{pix2pix} use a conditional GAN that takes a source image as input instead of the standard noise vector from original GANs to learn a mapping from said input to a target output image. A limitation of this approach is the need to have paired images from different distributions. In an effor to alleviate this constraint, Zhu et al. \cite{cyclegan} presented a similar but more flexible framework called CycleGAN that learns a mapping from distribution $X$ to $Y$ without the need for paired inputs. To ensure such "unsupervised" mapping, besides the common adversarial loss, they add a cycle consistency loss with the premise that a correctly learned inverse mapping should translate the generated image back to its original distribution. Similar to this work are DiscoGAN \cite{kim2017learning} and DualGAN \cite{yi2017dualgan} but with different loss functions. Some other unsupervised work assume a shared latent space between source and target domains. For instance, Liu et al. proposed unsupervised image-to-image translation (UNIT) \cite{liu2017unsupervised} building upon coupled GANs \cite{liu2016coupled} which enforce source and target generators and discriminators to share parameters in low-level layers assuming the existence of a shared latent space between two such distributions. Huang et al. extended this framework to a multi-modal setting in MUNIT \cite{huang2018multimodal}, using two latent representations for style and content. 

\textbf{Attention-based image-to-image translation.} The effectiveness of attention mechanisms on language models made researchers look for similar attention algorithms for vision tasks. However, due to the quadratic cost of the attention activations over the input sequence, which makes it undesirable for high resolution images, attention has been interpreted loosely in image-to-image translation. For instance, Tang et al. \cite{tang2019attentiongan} generate a series of foreground, background and content masks to guide the translation process, in a slightly more complex model than previous work. Mejjati et al. \cite{NEURIPS2018_mejjati} add two more attention networks that learn foreground and background attention maps and remove instance normalization from the discriminator. In SPA-GAN \cite{emami2020spa}, the discriminator learns an attention map that is fed to the generator to help focus the most discriminative object parts, but it is not composed by different masks.   

Motivated by the important role of attention mechanisms and advocating for a simpler model, we propose an efficient alternative for image-to-image translation based on weight-tied MLPs, which aims to alleviate the problem of ignored long-range dependencies. 

\section{MixerGAN}
We show that mixer blocks offer an alternative method of performing unpaired image-to-image translation that accounts for global relations not possible with vanilla convolutional blocks and in a computationally cheaper way than transformer blocks.

Formally, our goal is to train two mixer-based generators $G : X \to Y$ and $F : Y\to X$ that ``translate'' images between two distributions. We accomplish this in the same way as the original CycleGAN by also simultaneously training two discriminators $D_X$ and $D_Y$ to distinguish between real and generated images.

\subsection{Training Objective}
For a more stable optimization, we employ the LS-GAN formulation \cite{lsgan} such that our generators $G$ and $F$ are then trained to minimize
\begin{equation}
\begin{split}
    \mathcal{L}_G &= \mathbb{E}_{x\sim X}[(D_Y(G(x))-1)^2]\\
    \\
    \mathcal{L}_F &= \mathbb{E}_{y\sim Y}[(D_X(F(y))-1)^2]
\end{split}
\end{equation}
Our discriminators $D_X$ and $D_Y$ are then simultaneously trained to minimize
\begin{equation}
\begin{split}
    \mathcal{L}_{D_X} =& \mathbb{E}_{x\sim X}[(D_X(x) - 1)^2]\\
     + &\mathbb{E}_{y\sim Y}[D_X(F(y))^2]\\
    \\
    \mathcal{L}_{D_Y} =& \mathbb{E}_{y\sim Y}[(D_Y(y) - 1)^2]\\
     + &\mathbb{E}_{x\sim X}[D_Y(G(x))^2]
\end{split}
\end{equation}

In accordance with the CycleGAN \cite{cyclegan} formulation, we also regularize the cycle consistency loss between the two generators:
\begin{equation}
\begin{split}
    \mathcal{L}_{cyc} =& \mathbb{E}_{x\sim X}[|F(G(x)) - x|]\\
     +& \mathbb{E}_{y\sim Y}[|G(F(y)) - y|]
\end{split}
\end{equation}
The weighted cycle consistency loss is then added to both generator losses to encourage an invertible translation.

\subsection{Model Architecture}
\label{model}
\begin{figure}
    \centering
    \includegraphics[scale=0.3]{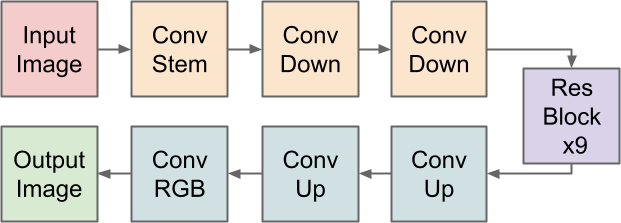}\\
    \vspace{5mm}
    CycleGAN Generator
    \vspace{10mm}
    \\
    \includegraphics[scale=0.3]{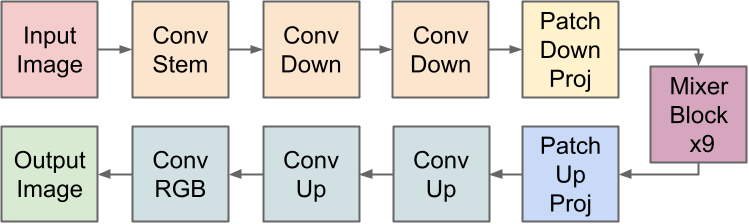}\\
    \vspace{5mm}
    MixerGAN Generator
    \caption{(Top): The CycleGAN architecture consists of 9 isotropic residual convolution blocks book-ended by down and up-sampling convolutions. (Bottom): In the MixerGAN architecture, we replace the residual blocks with MLP-Mixer blocks between down and up patch projections.}
    \label{fig:mixer_arch}
\end{figure}
For comparison's sake, we experiment with networks of similar size to those used in the original CycleGAN paper \cite{cyclegan}. While the mixer blocks have more parameters than the convolutional blocks, our mixer models still have the same number of non-linearities and a comparable---if not lower---dimensional latent space.

In accordance with the original CycleGAN, our mixer-based generator consists of a single-layer convolutional stem followed by two strided convolutional layers for learned down-sampling and a linear patch down-projection step. The patch down-projection step is effectively a convolution where the kernel size and stride are equal to the patch size, resulting in an output representation where each pixel is a linear projection of a disjoint patch of the input. For all our experiments, we used a patch size of 8x8 and projected each patch to a token vector of length 256. The transforming part of the MixerGAN generator then consists of 9 isotropic mixer blocks (whereas the analogous part of the CycleGAN generator has 9 residual convolutional blocks). Lastly, the transformative section is followed by a linear patch up-projection step, two up-sampling transposed convolutions, and a final convolution to remap the representation to the source RGB space. For our discriminator, we employ a PatchGAN in accordance with the CycleGAN paper \cite{pix2pix}.

The mixer blocks themselves directly follow the formulation given in the MLP-Mixer paper \cite{mixer}. After the patch projection step, the tokens are stacked as rows such that the representation $\mathbf{X}$ is now only two-dimensional and is of shape $(tokens \times channels)$. The mixer block itself consists of two Multi-Layer Perceptrons with skip connections: one for token mixing and one for channel mixing. The token-mixing MLP acts on the \textit{columns} of the representation while the the channel-mixing MLP acts on its \textit{rows}. If we do not expand the hidden dimension of the MLPs, then the shape of our token-mixing weights $\mathbf{W}_1$ and $\mathbf{W}_2$ is $(tokens \times tokens)$. Likewise, the shape of our channel-mixing weights $\mathbf{W}_3$ and $\mathbf{W}_4$ is $(channels \times channels)$.

Formally, the MLP-mixer block on $\mathbf{X} \to \mathbf{Y}$ can be described by the following equations:
\begin{equation}
\begin{split}
    \mathbf{U} &= \mathbf{X} + \mathbf{W}_2 \mbox{GELU}(\mathbf{W}_1 \mbox{LayerNorm}(\mathbf{X}))\\
    \mathbf{Y} &= (\mathbf{U}^T + \mathbf{W}_4\mbox{GELU}(\mathbf{W}_3\mbox{LayerNorm}(\mathbf{U})^T))^T
\end{split}
\end{equation}
where GELU is the Gaussian Error Linear Unit \cite{gelu}. Note that our mixer blocks are isotropic such that the shape of the output is the same as the input: $(tokens \times channels)$. Each linear operator of the token-mixing MLP mixes assigns each output token as a linear combination of \emph{all} input tokens where the linear weights depend on the token's location. As such, the MLP-mixer, by definition, applies long-range interactions between pixels.

An overview of our MixerGAN generator and a comparison to that of CycleGAN can be found in Figure \ref{fig:mixer_arch}. Pytorch \cite{pytorch} implementations of our MixerGAN generator and the MLP-Mixer block itself can be found in Appendix \ref{code}.

\subsection{Computational Advantages}
\begin{table}
   %Some results extracted from Mejjati et al. \cite{NEURIPS2018_mejjati}}
  \centering
  \begin{tabular}{cc|cc}
    \toprule
    \multicolumn{4}{c}{Activation Memory Usage Per Sample (Floats)}\\
    \midrule
    \#tokens & \#channels & SA & TM\\
    \midrule
    16x16 & 128  & 98.3k & 32.8k\\
    16x16 & 512  & 196.6k & 131.1k\\
    32x32 & 128  & 1179.6k & 131.1k\\
    32x32 & 512  & 1572.9k & 524.3k\\
    64x64 & 128  & 17301.5k & 524.3k\\
    64x64 & 512  & 18874.4k & 2097.2k\\
    \bottomrule
  \end{tabular}
  \caption{Self-attention (SA) memory usage scales linearly with channels but quadratically with tokens while token-mixing (TM) memory only scales linearly with both channels and tokens. Self-attention memory usage additionally scales linearly with the number of heads (values in table for 1-headed self-attention).}
    \label{table:memory_usage}
\end{table}
\begin{table}
   %Some results extracted from Mejjati et al. \cite{NEURIPS2018_mejjati}}
  \centering
  \begin{tabular}{cc|cc}
    \toprule
    \multicolumn{4}{c}{Trainable Parameters}\\
    \midrule
    \#tokens & \#channels & SA & TM\\
    \midrule
    16x16 & 128 & 49.1k & 65.5k\\
    16x16 & 512 & 786.4k & 65.5k\\
    32x32 & 128 & 49.1k & 1048.6k\\
    32x32 & 512 & 786.4k & 1048.6k\\
    64x64 & 128 & 49.1k & 16777.2k\\
    64x64 & 512 & 786.4k & 16777.2k\\
    \bottomrule
  \end{tabular}
  \caption{Self-attention's (SA) parameter count scales quadratically strictly with channels while token-mixing's (TM) scales quadratically strictly with tokens. However, parameter count is static and does not scale with batch size, so we don't see the same issues as when scaling self-attention to higher batch sizes.}
  \label{table:params}
\end{table}
For this analysis, we consider all layers to be ``isotropic'' such that their inputs and outputs are of the same shape.

As noted in the seminal paper \cite{mixer}, the MLP-Mixer is, at its core, a convolutional neural network with very specific architectural hyper-parameters. As such, the MLP-Mixer can exploit the existing GPU architectures and implementations that allow convolution operations to be performed with extreme efficiency, whereas attention-based networks are currently throttled by the speed at which GPUs can perform the un-optimized attention operation. 

Furthermore, the MLP-Mixer and transformer blocks differ in their usage of memory. Both the MLP-mixer block and transformer block contain a channel-mixing MLP, but the transformer uses a self-attention module in lieu of the MLP-mixer's token-mixing MLP. Since both the transformer and MLP-mixer contain a channel-mixing MLP, we will focus on the transformer's self-attention operator and the mixer's token-mixing MLP for comparison. 

For a representation with $n$ tokens and $c$ channels, the self-attention operator of the transformer block has $\mathcal{O}(c^2)$ parameters while the token-mixing MLP of the mixer block has $\mathcal{O}(n^2)$ parameters (Table ~\ref{table:params}). However, the main memory sink of the self-attention block comes in the intermediate activations necessary for both forward and back-propagation. For a batch size $b$, the $h$-headed self-attention module uses $\mathcal{O}(hbn^2 + bnc)$ intermediate activation floats while the token-mixing MLP uses $\mathcal{O}(bnc)$. This extensive increase in activation memory comes from the token-mixing part of the attention mechanism. A separate $n\times n$ attention score matrix is calculated for each attention head \textit{per sample}. Thus, the memory usage term that is quadratic in $n$ scales linearly with both $b$ and $h$. It is this additional $hbn^2$ term in the memory usage of the self-attention module that makes transformer models prohibitive for domains with a large number of tokens, such as visual data. Table \ref{table:memory_usage} illustrates how quickly the memory usage of self-attention explodes with the number of tokens. 

A vanilla residual block consisting of two convolutional layers with kernels of size $k\times k$ would only have $\mathcal{O}(k^2c^2)$ parameters and use $\mathcal{O}(bnc)$ intermediate activation floats. Clearly, a vanilla residual block has the least memory usage and parameter count, but it lacks the capacity to account for long-range relationships between tokens as described above due to the small receptive field.

\iffalse
\begin{figure}
    \centering
    \includegraphics[width=1\linewidth]{fig/complexity.png}
    \caption{(Left) Self-Attention (SA) uses a fixed number of parameters agnostic of the number of tokens while the Token Mixer's (TM) number of parameters scales quadratically with the number of tokens. (Center) Conversely, Self-Attention's activation memory scales quadratically with the number of tokens while the Token-Mixer's scales linearly. (Right) Both Self-Attention and the Token Mixer's activation scales linearly with batch size, however Self-Attention's scales much harder since it is also quadratic in the number of tokens.}
    \label{fig:memory_batch}
\end{figure}
\fi

\section{Experiments}
\begin{figure*}[]
    \centering
    \begin{tabular}{c c c }
       \includegraphics[width=0.20\linewidth]{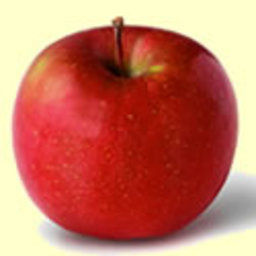}  &  \includegraphics[width=0.20\linewidth]{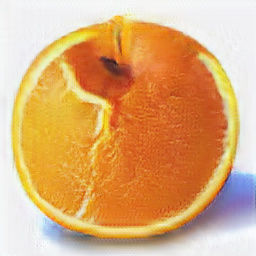} & \includegraphics[width=0.20\linewidth]{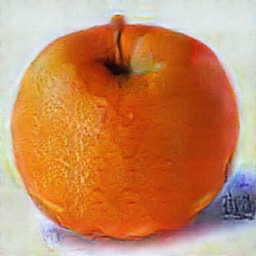}\\
       \includegraphics[width=0.20\linewidth]{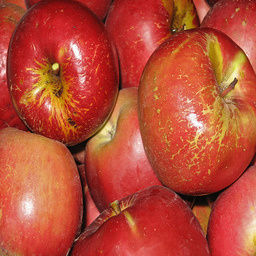}  &  \includegraphics[width=0.20\linewidth]{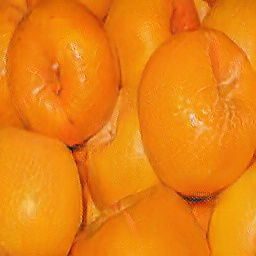} & \includegraphics[width=0.20\linewidth]{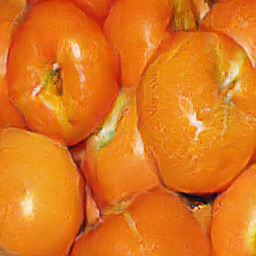}\\ 
       \includegraphics[width=0.20\linewidth]{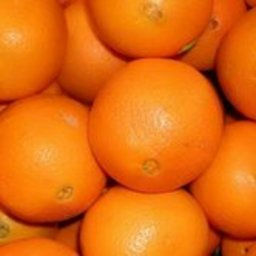}  &  \includegraphics[width=0.20\linewidth]{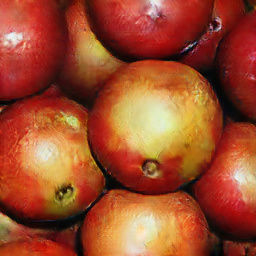} & \includegraphics[width=0.20\linewidth]{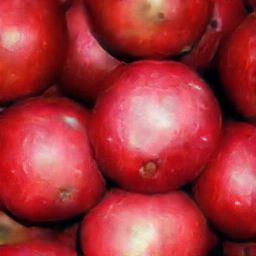}\\
       \includegraphics[width=0.20\linewidth]{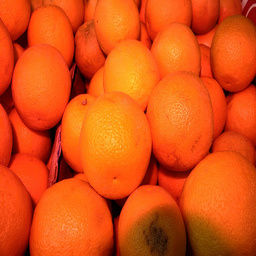}  &  \includegraphics[width=0.20\linewidth]{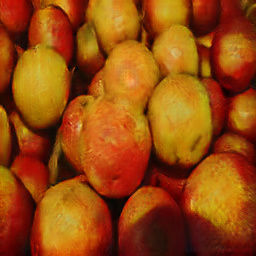} & \includegraphics[width=0.20\linewidth]{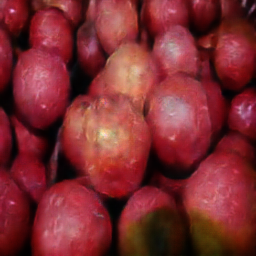}\\ 
       Source & CycleGAN & MixerGAN (Ours)
    \end{tabular}
    \caption{High-resolution results on translating between apples and oranges. Sources selected from those that gave ``Best Result'' on CycleGAN \cite{cyclegan} according to the authors. Despite the patch down and up-projection, MixerGAN is still able to generate coherent, high-resolution output.}
    \label{fig:fruit_results}
\end{figure*}
\begin{table*}[!th]
   %Some results extracted from Mejjati et al. \cite{NEURIPS2018_mejjati}}
  \centering
  \begin{tabular}{lcccc}
    \toprule
    \multicolumn{1}{c}{} & \multicolumn{4}{c}{KID ($\downarrow$)}\\
    \midrule
    Algorithm & A $\rightarrow$ O & O $\rightarrow$ A & H $\rightarrow$ Z & Z $\rightarrow$ H  \\
    \midrule
    DiscoGAN \cite{kim2017learning} & 18.34 $\pm$ 0.75 & 21.56 $\pm$ 0.80 & 13.68 $\pm$ 0.28 & 16.60 $\pm$ 0.50\\
    RA \cite{wang2017residual} & 12.75 $\pm$ 0.49 & 13.84 $\pm$ 0.78 & 10.16 $\pm$ 0.12 & 10.97 $\pm$ 0.26 \\
    DualGAN \cite{yi2017dualgan}& 13.04 $\pm$ 0.72 & 12.42 $\pm$ 0.88 & 10.38 $\pm$ 0.31 & 12.86 $\pm$ 0.50  \\
    UNIT \cite{liu2017unsupervised} & 11.68 $\pm$ 0.43 & 11.76 $\pm$ 0.51 & 11.22 $\pm$ 0.24 & 13.63 $\pm$ 0.34 \\
    CycleGAN \cite{cyclegan} & 8.48 $\pm$ 0.53 & 9.82 $\pm$ 0.51 & 10.25 $\pm$ 0.25 & 11.44 $\pm$ 0.38   \\
    CycleGAN (reproduced) & 8.42 $\pm$ 0.49 & 13.71 $\pm$ 0.62 & 3.53 $\pm$ 0.9 & 4.91 $\pm$ 0.67\\
    UA \cite{NEURIPS2018_mejjati} & 6.44 $\pm$ 0.69 & 5.32 $\pm$ 0.48 & 6.93 $\pm$ 0.27 & 8.87 $\pm$ 0.26  \\
    SPA-GAN \cite{emami2020spa} & 3.77 $\pm$ 0.32 & 2.38 $\pm$ 0.33 & 2.01 $\pm$ 0.13 & 2.19 $\pm$ 0.12  \\
    MixerGAN (Ours) & 10.23 $\pm$ 0.69 & 4.50 $\pm$ 0.44 & 6.81 $\pm$ 0.12 & 13.99 $\pm$ 0.25 \\
    \bottomrule
  \end{tabular}
  \caption{Kernel Inception Distance x100 $\pm$ std. x100 for different image translation algorithms. Lower is better. Abbreviations: A - Apple, O - Orange, H - Horse, Z - Zebra.}
  \label{kid_algos}
\end{table*}
We test MixerGAN on several datasets for unpaired image-to-image translation and include both quantitative and qualitative analysis of the results.

\paragraph{Training Details} To ensure a fair comparison of the effectiveness of our MLP-based model with respect to a CNN-based model, we follow the settings of CycleGAN \cite{cyclegan} as closely as possible. We leave downsampling and upsampling layers as the original model but change the core translation ResNet blocks of CycleGAN to our MLP-mixer blocks. That is, we have a total of 9 MLP-Mixer blocks, as explained in section \ref{model}. We use the original patch discriminator in CycleGAN. We use Adam optimizer with learning rate of 0.0003, keeping the same learning rate for the first 100 epochs and linearly decaying the learning rate to zero until the last iteration. All experiments are trained on an RTX A6000 GPU for 20,000 iterations with batch size of 16, patch size of 8x8, latent channel width of 256, and image input size of 256x256 pixels (unless otherwise specified). 

\paragraph{Baselines} We compare our results with CNN-based CycleGAN \cite{cyclegan}, DiscoGAN \cite{kim2017learning}, DualGAN \cite{yi2017dualgan}, UNIT \cite{liu2017unsupervised} and with attention-based models from Mejjati et al. \cite{NEURIPS2018_mejjati}, SPA-GAN \cite{emami2020spa}, and Residual Attention (RA) \cite{wang2017residual}. To ensure a fair comparison, we reproduce CycleGAN, and evaluate our method on Kernel Inception Distance (KID) \cite{binkowski2018kid} scores. While Fr\`echet Inception Distance (FID) is a more commonly used metric today, many of the models to which we are comparing only report KID. We unfortunately do not have the compute to reproduce all these other models in a reasonable amount of time to determine their FIDs, so we only report our comparisons in terms of KID. Since this is an exploratory work to show the feasibility of MLP-based image-to-image translation and not a claim of state-of-the-art performance, we maintain that the KID comparison is sufficient.

\paragraph{Datasets} We use the following datasets for image-to-image translation provided by Zhu et al. \cite{cyclegan}: Apple-to-Orange (A $\leftrightarrow$ O), containing 995/266 train/test apple images and 1019/248 train/test orange images; Horse-to-Zebra (H $\leftrightarrow$ Z), containing 1067/120 train/test horse images and 1334/140 train/test zebra images; and Yosemite Summer-to-Winter (S $\leftrightarrow$ W), containing 1231/309 train/test summer images, and 962/238 train/test winter images.
\iffalse
\paragraph{Ablations} We study the impact that channel width, patch size have in the results, as well as the addition of a perceptual loss in the discriminator. Table \ref{ablations} shows FID \cite{heusel2017fid} and KID \cite{binkowski2018kid} scores for the Apple to Orange dataset. Perceptual loss is calculated over feature maps, extracted from using a pretrained VGG-16 \cite{simonyan2014vgg}. We use VGG's \textit{conv 1\_1} to \textit{conv 3\_1} layers for the perceptual loss calculation. We found that at lower resolution (128x128) perceptual loss reduces the amount of generated artifacts in the output. However, when using 256x256 input size images, perceptual loss does not generate better results. The effect of the perceptual loss weight $\lambda$ is shown in table \ref{ablations}. We found that a weight higher than $0.001$ decreases performance, yielding worse generated outputs. 

Our Mixer-MLP takes in a sequence of patches. An optimal patch size depends on the input size. We found that a patch size of 4 gives the best results for images of size 128x128, whereas a patch size of 8 works better for 256x256 images. Generally, a depth of 256 works well for images of size 256. However, channel depth of 128 gives similar results in less training time. For image size of 256 and patch size of 8, training for 20,000 iterations, channel depth of 128 takes around 380 minutes, whereas a channel depth of 256 takes around 660 minutes. 
\fi

\section{Results}
\begin{figure*}[]
    \centering
    \begin{tabular}{c @{\hskip 0.05in} c @{\hskip 0.05in} c @{\hskip 0.2in} c @{\hskip 0.05in} c @{\hskip 0.05in} c}
    \\
    \\
       \includegraphics[width=0.15\linewidth]{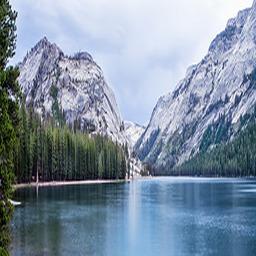}  &  \includegraphics[width=0.15\linewidth]{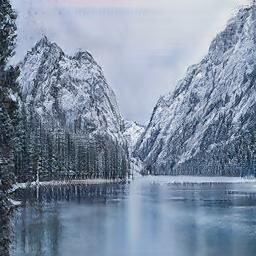} & \includegraphics[width=0.15\linewidth]{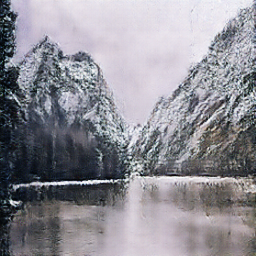} &
       \includegraphics[width=0.15\linewidth]{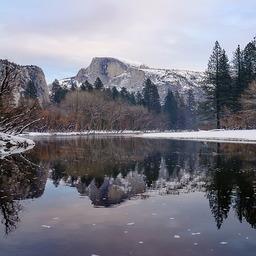}  &  \includegraphics[width=0.15\linewidth]{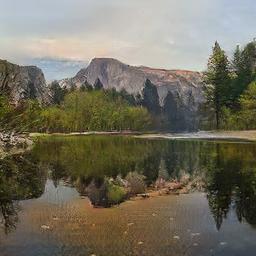} & \includegraphics[width=0.15\linewidth]{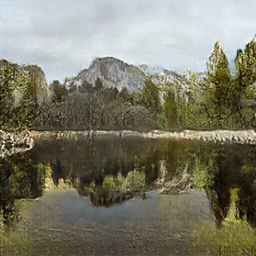} \\
       \includegraphics[width=0.15\linewidth]{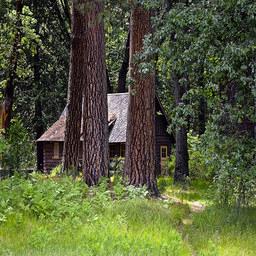}  &  \includegraphics[width=0.15\linewidth]{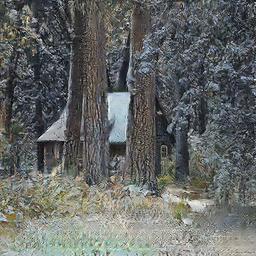} & \includegraphics[width=0.15\linewidth]{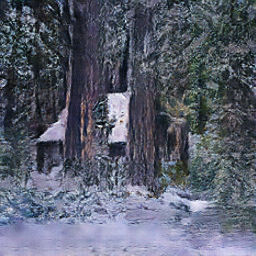} & 
       \includegraphics[width=0.15\linewidth]{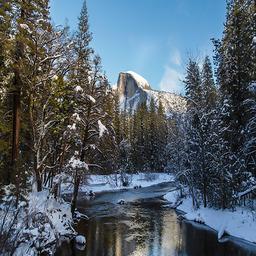}  &  \includegraphics[width=0.15\linewidth]{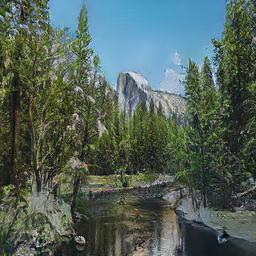} & \includegraphics[width=0.15\linewidth]{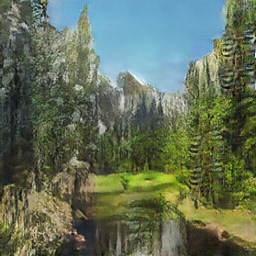}\\ 
        \includegraphics[width=0.15\linewidth]{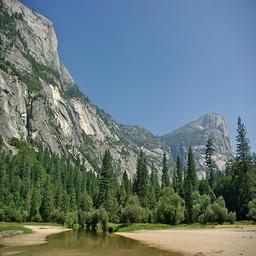}  &  \includegraphics[width=0.15\linewidth]{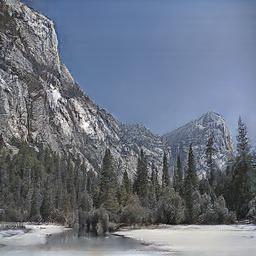} & \includegraphics[width=0.15\linewidth]{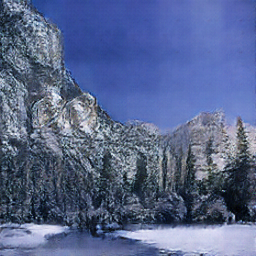} &
       \includegraphics[width=0.15\linewidth]{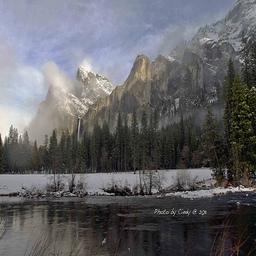}  &  \includegraphics[width=0.15\linewidth]{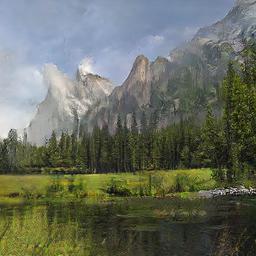} & \includegraphics[width=0.15\linewidth]{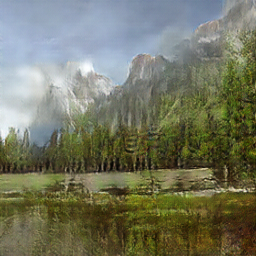} \\
        \includegraphics[width=0.15\linewidth]{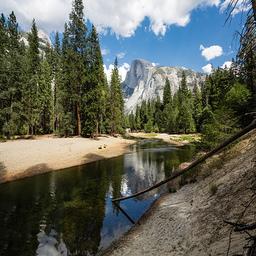}  &  \includegraphics[width=0.15\linewidth]{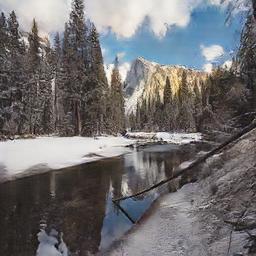} & \includegraphics[width=0.15\linewidth]{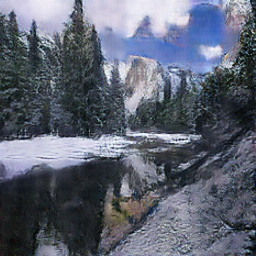} &
       \includegraphics[width=0.15\linewidth]{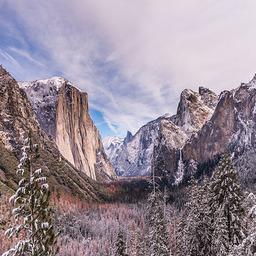}  &  \includegraphics[width=0.15\linewidth]{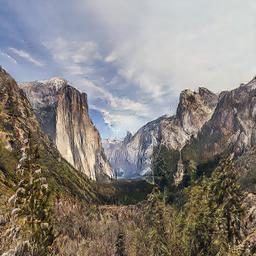} & \includegraphics[width=0.15\linewidth]{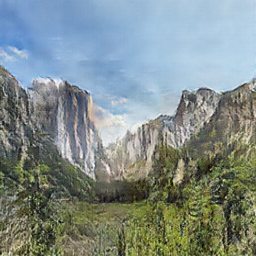} \\
       Source & CycleGAN & MixerGAN & Source & CycleGAN & MixerGAN\\
       \\
       \multicolumn{3}{c}{Summer to Winter} & \multicolumn{3}{c}{Winter to Summer}
    \end{tabular}
    \caption{Results on the Yosemite dataset. MixerGAN better captures the vibrant whites and purples of winter and blues and greens of summer while CycleGAN has a more muted pallet.}
    \label{fig:yosemite_results}
\end{figure*}

As previously stated, the objective of this work is not to achieve state-of-the-art results or out-perform any baselines. Instead, our purpose is to show that an MLP-based architecture can effectively perform image-to-image translation while accounting for long-range connections not possible with convolution and being much less computationally expensive than a transformer-based model. Adapting further improvements upon the CycleGAN architecture to MixerGAN, such as contrastive learning \cite{cut}, are left as the subject of future work.

\subsection{Quantitative Results}
For completion's sake, we include a quantitative comparison of MixerGAN's with other popular image-to-image translation algorithms.

To quantitatively evaluate our model, we use Kernel Inception Distance (KID) \cite{binkowski2018kid}. KID computes the squared maximum mean discrepancy between real and fake images using feature representations, extracted from the Inception network architecture \cite{szegedy2016inception}. Similar to Fr\'echet Inception Distance (FID) metric \cite{heusel2017fid}, and even though KID is not bounded, a lower score implies the generated distribution is closer to the real distribution such that it is visually more similar. We compute KID scores on unidirectional mappings, from $X$ to $Y$ and vice-versa. 

On all tasks, MixerGAN performs within the same KID range as the other models, as seen in Table \ref{kid_algos}. Compared directly to CycleGAN \cite{cyclegan}, MixerGAN significantly out-performs both the authors' originally reported and our re-produced KID values on the Orange-to-Apple task while remaining in the same ballpark for tasks on which it scored worse. Recall, our goal is \textbf{not} to outperform current methods, but to instead show the feasibility of MLP-based models as a vehicle for image-to-image translation. The fact that MixerGAN scores quantitatively in the same range as current models supports this idea.

\iffalse
\begin{table*}
  \centering
  \begin{tabular}{lcccc}
    \toprule
    \multicolumn{1}{c}{} & \multicolumn{2}{c}{FID} & \multicolumn{2}{c}{KID} \\
    \midrule
    Algorithm & A $\rightarrow$ O & O $\rightarrow$ A  & A $\rightarrow$ O & O $\rightarrow$ A \\
    \midrule
    MixerGAN C256 P8 $\lambda$p 0.001 & 185.72 & 141.77  & 10.92 $\pm$ 0.11 & 5.38 $\pm$ 0.63    \\
    MixerGAN C128 P8 $\lambda$p 0.001 & 189.36 & 135.00  & 11.41 $\pm$ 0.01 & 5.04 $\pm$ 0.19   \\
    MixerGAN C256 P8 $\lambda$p 0.0005 & 184.67 &  149.54  & 10.57 $\pm$ 0.33 & 6.17 $\pm$ 0.52 \\
    MixerGAN C128 P8 $\lambda$p 0.0005 & 190.74 &  135.46  & 11.79 $\pm$ 0.17 & 4.86 $\pm$ 0.32  \\
    MixerGAN C256 P8 $\lambda$p 0 & 178.73 & \textbf{127.92} & 10.63 $\pm$ 0.07 & \textbf{4.50} $\pm$ \textbf{0.24}\\
    MixerGAN C128 P8 $\lambda$p 0  & \textbf{177.94} & 140.09  & \textbf{10.27} $\pm$ \textbf{0.19} & 5.61 $\pm$ 0.12  \\
    \bottomrule
  \end{tabular}
  \caption{Ablation Study. Fr\'echet Inception Distance and Kernel Inception Distance x100 $\pm$ std. x100. Input image size 256x256 pixels}
\label{ablations}
\end{table*}
\fi

\subsection{Qualitative Results}

For our qualitative results, we compare the output of our model to that of CycleGAN \cite{cyclegan} on images that the CycleGAN authors note gave them the best results.

\paragraph{Comparing Apples and Oranges}In Figure \ref{fig:fruit_results}, we present our results in large resolution so the reader can appreciate that our MixerGAN can produce images of the same fidelity as CycleGAN. For the sake of space, we do not repeat this for our other datasets. Examining Figure \ref{fig:fruit_results}, we see that not only can MixerGAN create images of similar quality as CycleGAN, but it even improves upon it in some cases. In the first row, we see that CycleGAN generated an unrealistic combination of the inside and outside of an orange while MixerGAN generated a coherent-looking orange. In the third row, we see that CycleGAN retained the orange's texture in the generated apple, especially near the specular reflections. In contrast, MixerGAN correctly synthesizes the typical apple texture, including around the specualar reflections.

\paragraph{Yosemite Dataset}
Our model sees an encouraging level of success on the Yosemite dataset. In Figure \ref{fig:yosemite_results}, we see the results of MixerGAN compared to those of CycleGAN \cite{cyclegan} on some of the ``best'' images as selected by the CycleGAN authors. From a purely subjective standpoint, MixerGAN seems to do a better job achieving the vibrant colors of the respective seasons. Specifically, we see in the synthetic winter images that MixerGAN generates much whiter snow and more purple skies whereas CycleGAN tends to generate a blue-gray haze. In the synthetic summer images, MixerGAN generates significantly more saturated blues and greens while CycleGAN yields a more muted palette. 

\paragraph{Horses to Zebras}
\begin{figure*}[]
    \centering
    \begin{tabular}{c @{\hskip 0.05in} c @{\hskip 0.05in} c @{\hskip 0.2in} c @{\hskip 0.05in} c @{\hskip 0.05in} c}
    \\
    \\
       \includegraphics[width=0.15\linewidth]{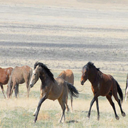}  &  \includegraphics[width=0.15\linewidth]{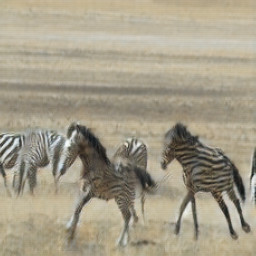} & \includegraphics[width=0.15\linewidth]{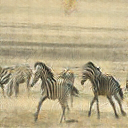} &
       \includegraphics[width=0.15\linewidth]{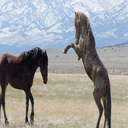}  &  \includegraphics[width=0.15\linewidth]{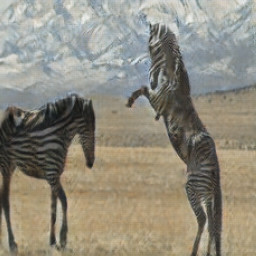} & \includegraphics[width=0.15\linewidth]{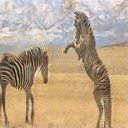} \\
       \includegraphics[width=0.15\linewidth]{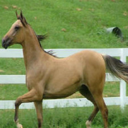}  &  \includegraphics[width=0.15\linewidth]{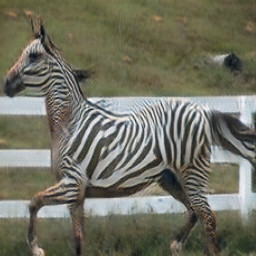} & \includegraphics[width=0.15\linewidth]{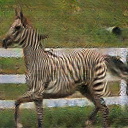} & 
       \includegraphics[width=0.15\linewidth]{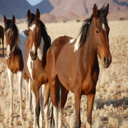}  &  \includegraphics[width=0.15\linewidth]{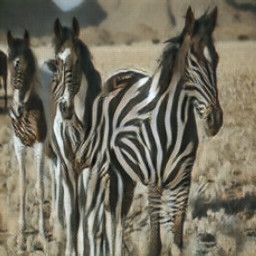} & \includegraphics[width=0.15\linewidth]{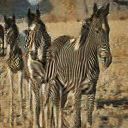}\\ 
       Source & CycleGAN & MixerGAN & Source & CycleGAN & MixerGAN\\
       \\
       \multicolumn{6}{c}{Horse to Zebra}
    \end{tabular}
    \caption{By using a target resolution of 128x128 instead of 256x256, we were able to obtain high-fidelity Horse to Zebra translations. However, the Zebra to Horse translations (not pictured) were still low quality.}
    \label{fig:zebras}
\end{figure*}
The Horse to Zebra dataset posed a large challenge for our model. When run on 256x256 images, the results were full of patch artifacts. We hypothesize that this is due to the previously mentioned information loss at the patch projection step and could be placated by using more channels in the latent representation. We then tried our model on 128x128 images and reached much better results for Horse to Zebra as seen in Figure \ref{fig:zebras} but still not for Zebra to Horse.

\subsection{Failure Modes and Limitations}
\begin{figure}[]
    \centering
    \begin{tabular}{c c c}
        \includegraphics[width=0.28\linewidth]{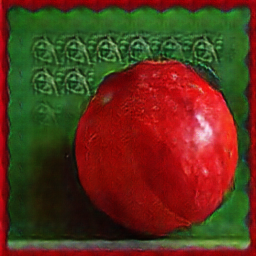} &  \includegraphics[width=0.28\linewidth]{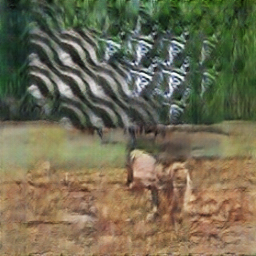}&
        \includegraphics[width=0.28\linewidth]{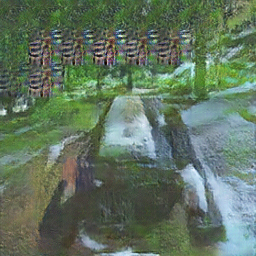} 
    \end{tabular}
    \caption{MixerGAN's most common failure mode is the repetition of an identical patch artifact. We hypothesize this would be less severe in a MixerGAN with a larger number of latent channels such that less information is destroyed at the patch projection step.}
    \label{fig:fail}
\end{figure}
The largest limiting factor of the MixerGAN model seems to come from the patch projection step. If our patch size is $(p\times p)$, then the spatial resolution is effectively reduced by a factor of $p^2$. If we only multiply the number of channels by 2 at the projection step, then we only retain $2/p^2$ of our activation dimensionality. In our experiments, we used a patch size of $(8 \times 8)$, so depending on the rank of our data at that point in the network and its effective compressibility, we could potentially lose almost 97\% of our information through the patch projection. Given that neighboring pixels are typically highly correlated, the practical information loss is likely much less, but it still limits the translating capacity of our generator.

Given the limitations of the patch projection step, we sometimes find ``patch'' artifacts in our generated images as seen in Figure \ref{fig:fail}. We hypothesize that a larger number of channels after our patch projection step would help alleviate this artifacting issue by stymieing the information loss at that point. A less under-complete representation at the translating phase would ideally result in less compression and a more expressive model overall. Unfortunately, with our modest amount of computing power, we were not able to experiment with wider networks at this time.

\section{Conclusions}
The recent proposal of the MLP-Mixer model for vision showed that multi-layer perceptrons can still be effective for visual classification tasks in the modern age of deep learning. In this work, we have shown that the MLP-Mixer is also an effective architecture for generative models, specifically unpaired image-to-image translation. We hypothesize that increasing the number of channels in the latent space would reduce any patch artifacting and  hope to obtain computing resources to investigate this in the future.

Like all image synthesis applications, MixerGAN has the potential to be used for potentially malicious goals, such as deepfakes \cite{deepfake}. As such, synthetic image detection continues to be an important field of its own. However, this should not deter us from continuing research in image synthesis, as the best way to combat a threat is to thoroughly understand it.

After decades of convolutional neural networks (and more recently, transformer networks) dominating the field of computer vision, it is remarkable that a simple sequence of weight-tied MLPs can effectively perform the same tasks. Now that we have shown that the MLP-Mixer succeeds at generative objectives, the door is open to refine this technique and extend MLP-based architectures to further image synthesis tasks.

{\small
\bibliographystyle{ieee_fullname}
\bibliography{wacv}
}
\clearpage
\appendix
\onecolumn
\section{Mixer Block and Generator Code}
\label{code}
\begin{verbatim}
class CycleGeneratorMixer(nn.Module):

    def __init__(self, patch_dim, image_size=256, embed_dim=256, transform_layers=9, 
                    patch_size=8):
        super(CycleGeneratorMixer, self).__init__()

        # stem
        model = [
            nn.Conv2d(in_channels=3, out_channels=embed_dim//4, kernel_size=7, 
                        padding=3, padding_mode='reflect'),
            nn.InstanceNorm2d(embed_dim//4),
            nn.ReLU(True),
            # PrintLayer(name='stem')
        ]

        # downsampling
        model += [
            nn.Conv2d(in_channels=embed_dim//4, out_channels=embed_dim//2, 
                        kernel_size=3, stride=2, padding=1),
            nn.InstanceNorm2d(embed_dim//2),
            nn.ReLU(True),
            nn.Conv2d(in_channels=embed_dim//2, out_channels=embed_dim, kernel_size=3, 
                        stride=2, padding=1),
            nn.InstanceNorm2d(embed_dim),
            nn.ReLU(True),
        ]

        # linear down-projection
        model += [nn.Conv2d(in_channels=embed_dim, out_channels=embed_dim, 
                    kernel_size=patch_size, stride=patch_size)]
       
        # reshape to (tokens, channels)
        model += [View((embed_dim, patch_dim))]

        # transformation
        model += [MixerBlock(embed_dim=embed_dim, patch_dim=patch_dim) 
                    for _ in range(transform_layers)]

        # reshape to (c, h, w)
        model += [View((embed_dim, image_size//4//patch_size, 
                    image_size//4//patch_size))]

        # linear up-projection
        model += [nn.ConvTranspose2d(in_channels=embed_dim, out_channels=embed_dim, 
                    kernel_size=patch_size, stride=patch_size)]
        
        # upsampling
        model += [
            nn.ConvTranspose2d(in_channels=embed_dim, out_channels=embed_dim//2, 
                                kernel_size=3, stride=2, padding=1, output_padding=1),
            nn.InstanceNorm2d(embed_dim//2),
            nn.ReLU(True),

            nn.ConvTranspose2d(in_channels=embed_dim//2, out_channels=embed_dim//4, 
                                kernel_size=3, stride=2, padding=1, output_padding=1),
            nn.InstanceNorm2d(embed_dim//4),
            nn.ReLU(True),
        ]


        # to RGB
        model += [nn.Conv2d(in_channels=embed_dim//4, out_channels=3, kernel_size=7, 
                    padding=3, padding_mode='reflect')]

        model += [nn.Tanh()]

        self.model = nn.Sequential(*model)

    def forward(self, x):
        return self.model(x)





class MixerBlock(nn.Module):
    def __init__(self, embed_dim, patch_dim):
        super(MixerBlock, self).__init__()
        self.ln1 = nn.LayerNorm([embed_dim, patch_dim])

        self.dense1 = nn.Linear(in_features=patch_dim, out_features=patch_dim, 
                                    bias=False)
        self.gelu1 = nn.GELU()
        self.dense2 = nn.Linear(in_features=patch_dim, out_features=patch_dim, 
                                    bias=False)

        self.ln2 = nn.LayerNorm([embed_dim, patch_dim])

        # using conv1d with kernel_size=1 is like applying a 
        # linear layer to the channel dim
        self.conv1 = nn.Conv1d(in_channels=embed_dim, out_channels=embed_dim, 
                                    kernel_size=1, bias=False)
        self.gelu2 = nn.GELU()
        self.conv2 = nn.Conv1d(in_channels=embed_dim, out_channels=embed_dim, 
                                    kernel_size=1, bias=False)

    def forward(self, x):
        # token-mixing mlp
        skip = x
        x = self.ln1(x)
        x = self.dense1(x)
        x = self.gelu1(x)
        x = self.dense2(x)
        x = x + skip

        # channel-mixing mlp
        skip = x
        x = self.ln2(x)
        x = self.conv1(x)
        x = self.gelu2(x)
        x = self.conv2(x)
        x = x + skip

        return x
\end{verbatim}
\end{document}